\documentclass{./class_files/llncs}
\usepackage{makeidx}  
\usepackage{booktabs}
\usepackage{graphicx}
\usepackage{amsfonts}
\usepackage{mathtools}
\usepackage{todonotes}
\usepackage[utf8]{inputenc}
\usepackage{subfigure}
\usepackage{multirow}
\usepackage{booktabs}
\newcommand{\citep}[1]{\cite{#1}}

\newcommand{\citet}[1]{\cite{#1}}

\newcommand\given[1][]{\:#1\vert\:}

\makeatletter
\def\blfootnote{\gdef\@thefnmark{}\@footnotetext}
\makeatother

\usepackage{hyperref}
\begin{document}
\mainmatter              

\title{One Network To Segment Them All:\\A General, Lightweight System for Accurate 3D Medical Image Segmentation}
\titlerunning{One Network To Segment Them All}

\blfootnote{Preprint, final version published in \emph{Medical Image Computing and Computer Assisted Intervention (MICCAI)}, LNCS 11765, pp.~30--38, Springer, 2019. The final authenticated publication is available online at \url{https://doi.org/10.1007/978-3-030-32245-8_4}}

\author{Mathias Perslev\inst{1} \and
		Erik Bjørnager Dam\inst{1,2} \and
		Akshay Pai\inst{1,2} \and
        Christian Igel\inst{1}}

\institute{Department of Computer Science, University of Copenhagen, \email{map@di.ku.dk}
\and Cerebriu A/S, Copenhagen, Denmark}

\authorrunning{M. Perslev et al.} 

\maketitle              

\begin{abstract}
Many recent medical segmentation systems rely on powerful deep learning models to solve highly specific tasks. To maximize  performance, it is standard practice to evaluate numerous pipelines with varying model topologies, optimization parameters, pre- \& postprocessing steps, and even model cascades. It is often not clear how the resulting pipeline transfers to different tasks.

We propose a simple and thoroughly evaluated deep learning framework for segmentation of arbitrary medical image volumes. The system requires no task-specific information, no human interaction and is based on a fixed model topology and a fixed hyperparameter set, eliminating the process of model selection and its inherent tendency to cause method-level over-fitting. The system is available in open source and does not require deep learning expertise to use. Without task-specific modifications, the system performed better than or similar to  highly specialized deep learning methods across 3 separate segmentation tasks. In addition, it ranked 5-th and 6-th in the first and second round of the 2018 Medical Segmentation Decathlon comprising another 10 tasks. 

The system relies on \emph{multi-planar} data augmentation which facilitates the application of a single  2D architecture based on the familiar U-Net. Multi-planar training combines the parameter efficiency of a 2D fully convolutional neural network with a systematic train- and test-time augmentation scheme, which allows the 2D model to learn a representation of the 3D image volume that fosters generalization.

\end{abstract}

\section{Introduction}
More and more systems for medical image segmentation rely on deep learning (DL). However, most  publications on this topic  report performance improvements for  a particular segmentation task and imaging modality and use 
a specialized processing pipeline adapted through hyperparameter tuning. 
This makes it difficult to generalize the obtained results and bears the risk that the reported findings are artifacts. 
In line with the idea behind the 2018 Medical Segmentation Decathlon (MSD)\footnote{\url{http://medicaldecathlon.com}} \citep{msd},
a challenge evaluating the generalisability of machine learning based segmentation algorithms, we argue that new segmentation systems should  be evaluated across many different data cohorts and maybe even tasks. 
This reduces the risk of unintentional method overfitting and may help to gain more general insights about, for example,  superior model architectures and learning methods for particular problem classes. This does not only contribute to our basic understanding of the segmentation algorithms, but also to  the  clinical acceptance and applicability of the systems -- even if the  generality could come at the cost of not reaching state-of-the-art performance on each individual cohort or task.

A DL segmentation framework that works across a wide range of tasks and in which the individual components and hyperparameters are sufficiently understood allows to automate the task-specific adaptations. This is a prerequisite for being useful for practitioners who are not experts in DL. 
Big compute clusters offer a way to design systems that provide accurate segmentations for a variety of  tasks and do not require tuning by DL experts. If compute resources are not limited, automatic model and hyperparameter selection can be implemented. Given  new training data, the systems tests a large variety of segmentation algorithms and, for each algorithm, explores the space of the required hyperparameters.
While this approach may produce powerful systems, and was employed to variable extents by top-performing MSD submissions, we argue that it has crucial drawbacks. First, it comes with a risk of automated method overfitting, even if the data is handled carefully. Second, the approach may be prohibitive in clinical practice (and for many scientific institutions) when there is simply no access to sufficient (data regulations compliant) compute resources.

This paper presents an open-source system for medical volume segmentation that addresses all the issues outlined above.
It relies on a single neural network of fixed architecture that \textbf{1)} showed very good performance across a variety of diverse segmentation tasks, \textbf{2)} can be trained efficiently  without DL expert knowledge, large amounts of data, and compute clusters, and \textbf{3)} does not need large resources  when deployed.
The system architecture is a 2D U-Net \citep{unet,dental} variant. The decisive feature of our approach lies in extensive data augmentation, in particular by rotating the input volume before presenting slices to the fully convolutional network. Because of the latter, we refer to our approach as \emph{multi-planar} U-Net training (\emph{MPUnet}).
We present a thorough evaluation of our system on a total of 13 different 3D segmentation tasks, including 10 from MSD, on which it obtains high accuracies -- often reaching state-of-the-art performance from even highly specialized DL-based methods.

\section{Method}




At the heart of our system lies a 2D U-net \citep{unet} modified slightly to \textbf{1)} include batch normalization layers \citep{bn} intervening each double convolution- and up-convolution block and \textbf{2)} use nearest-neighbor up-sampling followed by convolution to implement up-convolutions \cite{nn_upsampling}.
Basic network topology and hyperparameters can bet set to their default choices as done in all experiments in this paper, see Table \ref{supplementary:hyperparams} in the supplementary material for an overview.
Compared to \citep{unet}, the number of filters has been increased by a factor of $\sqrt{2}$, see supplementary Table \ref{supplementary:topology} for details.
As a result, the model has $\approx 62$ million parameters. While one would assume that the size of the model is a crucial hyperparameter, we kept the model architecture  the same for all tasks. For each task, only the filters in the first layer were resized according to the number $C$ of input channels and the number of output units was set to the number  of classes $K$.

\begin{figure}[t!]
	\centering
    \makebox[\textwidth][c]{
    	\includegraphics[width=1\linewidth]{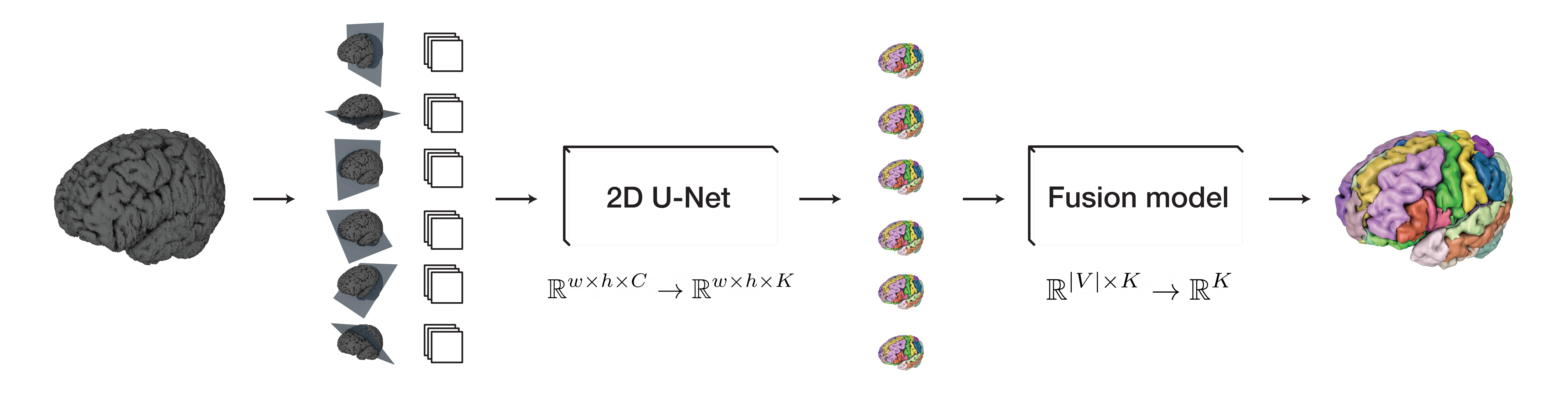}
    }
    \vspace{-3ex}\caption{Model overview. In the inference phase, the input volume (left) is sampled on 2D isotropic grids along multiple view axes. The model predicts a full volume along each axis and maps the predictions into the original image space. A fusion model combines the 6 proposed segmentation volumes into a single final segmentation.}
    \label{fig:model}
\end{figure}

The decisive feature of our multi-planar U-Net training (\emph{MPUnet}) is the generation of the inputs at training and test time, which is done by sampling from multiple planes of random orientation spanning the image volume. That is,  the network must learn to segment the input seen from  different views, see Fig.~\ref{fig:model}.

The model $f(x; \theta)$ takes as input multi-channel 2D image slices of size $w\times h$, $x \in \mathbb{R}^{w \times h \times C}$, and outputs a probabilistic segmentation map $P \in \mathbb{R}^{w \times h \times K}$ for  $K$ classes. Prior to training we define a set $V = \{ v_1, v_2, ..., v_i \}$ of $i$ randomly sampled unit vectors in $\mathbb{R}^3$. The set defines the axes through the image volume along which we sample 2D inputs to the model, visualized in Fig.~\ref{fig:views}. We re-sample the set $V$ until all pairs of vectors 
have an angle of at least $60 \deg$ between them.
A sampled set of planar axes is shown in Fig.~\ref{fig:viewsA}. Note that the model could also be fit using a set of fixed, predefined planes, but we found no performance gain in doing so, even if the fixed set included the standard planes. We use $i=6$ for all reported evaluations. This number was chosen based on prior experiments in which we observed monotonically improving performance with the inclusion of additional planes and $i=6$ providing a good balance between accuracy and computation, see supplementary Table \ref{supplementary:number_of_planes}.

During training, the model is provided batches of images randomly sampled from the $i$ planes in $V$ without supplying information about the corresponding axis. During inference, the model predicts along each plane producing a set of $i$ segmentation volumes $\mathbf{P} = \{ P_{v} \in \mathbb{R}^{w \times h \times d \times K} \given v \in V \}$. Each $P_{v}$ is mapped to the input image space to obtain point correspondence by assigning to each voxel in the input image the value of its nearest predicted point in $P_{v}$. Distances are computed in physical coordinates.

At test-time, the learned invariance to orientation is exploited by segmenting the entire volume from each view. This results in several candidate segmentations for each subject, which are combined by a linear fusion model, see  Fig.~\ref{fig:model}. 
We map $\mathbf{P}$ to a single probabilistic segmentation 
by a weighted sum of the per-class and per-view softmax-scores.
For all $w \cdot h \cdot d$ voxels $x$ in $\mathbf{P}$ and each class $k \in \{ 1, ..., K\}$, the \emph{fusion model} $f_{\text{fusion}} : \mathbb{R}^{|V| \times K} \rightarrow \mathbb{R}^{K}$ calculates
%
	$z(x)_k = \sum_{n=1}^{|V|} W_{n, k} \cdot p_{n, x, k} + \beta_k$.
Here $p_{n, x, k}$ denotes the probability 
of class $k$ at voxel $x$ as predicted by segmentation $P_n$.
The $W \in \mathbb{R}^{|V| \times K}$ weighs the probabilities of each class as predicted from each view and $\beta \in \mathbb{R}^{K}$ are bias parameters, which can adjust the  overall tendency to predict a given class. The parameters of $f_{\text{fusion}}$ are learned from the validation data. The model scales the predictions according to which views do well on each class, motivated by the fact that different target classes may appear in different shapes and levels of recognizability when seen from the different directions in $V$.


\newlength{\mylength}
\setlength{\mylength}{0.33\linewidth}
\begin{figure}[t!]
\centering
\makebox[.75\textwidth][c]{
  \subfigure[]{\label{fig:viewsA}\includegraphics[width=\mylength]{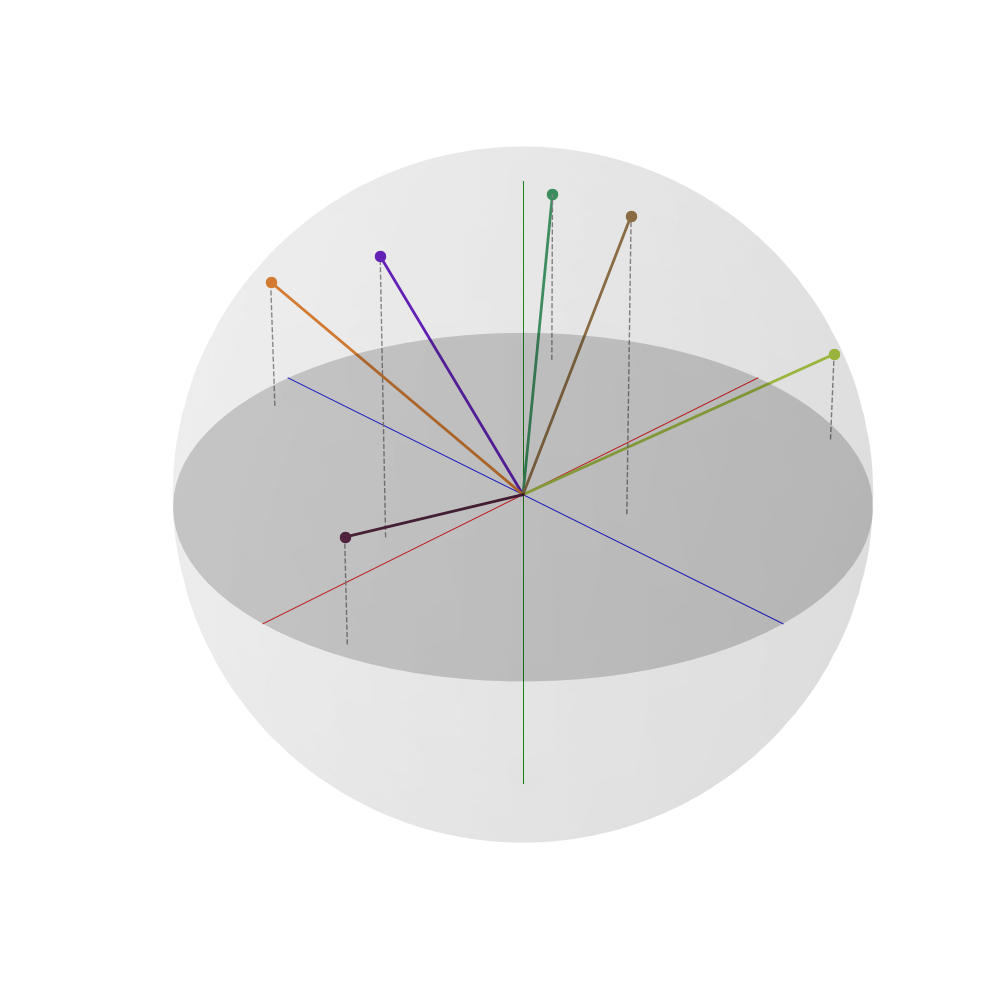}}
  \subfigure[]{\label{fig:viewsB}\includegraphics[width=\mylength]{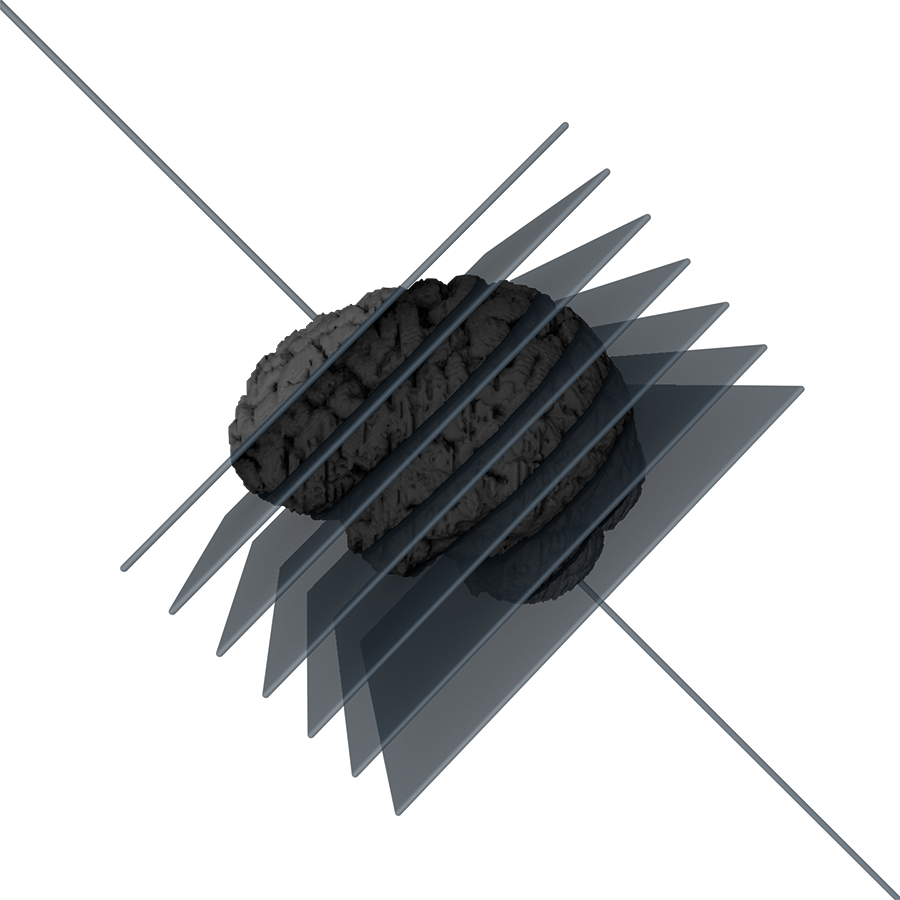}}
  \subfigure[]{\label{fig:viewsC}\includegraphics[width=\mylength]{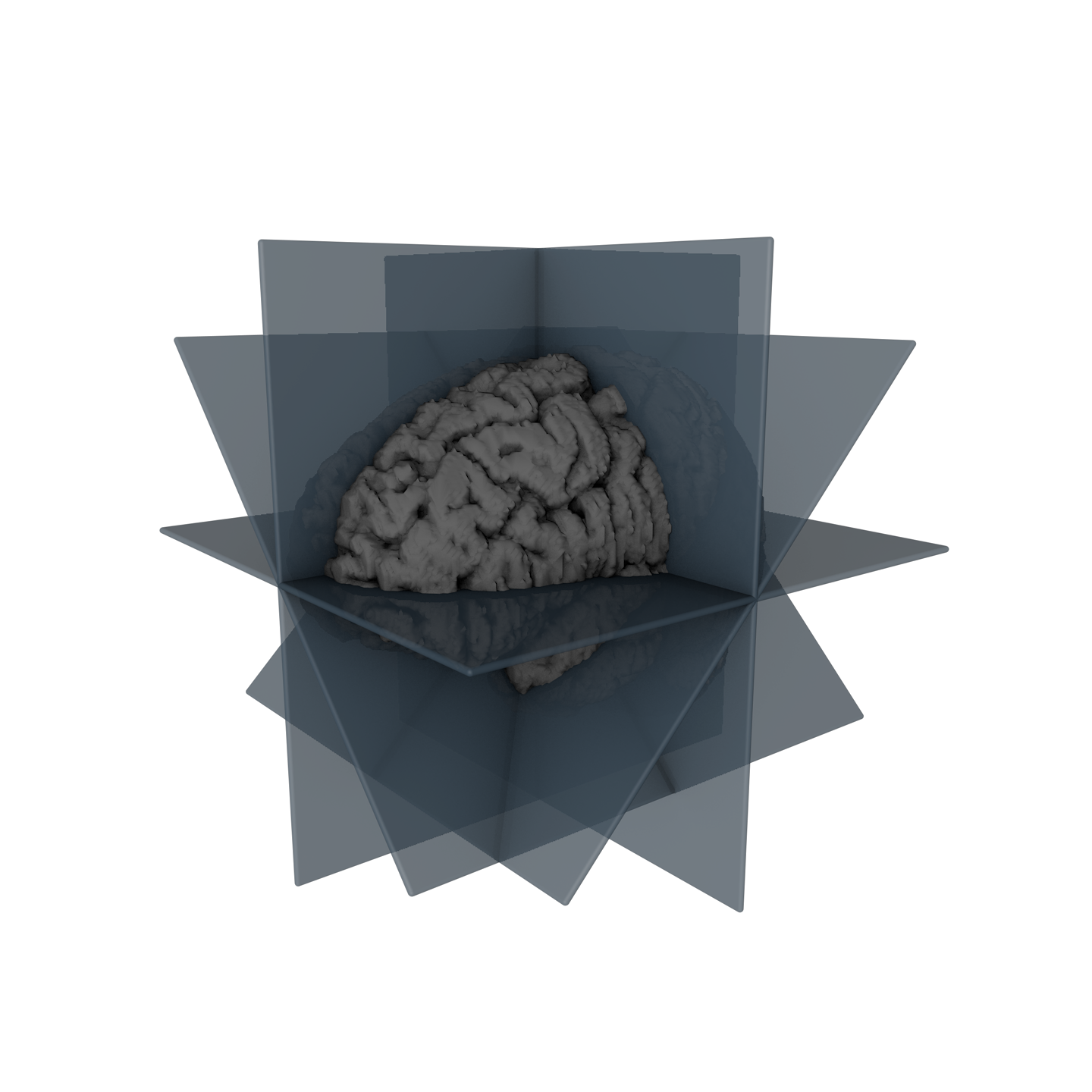}}
  \vspace{-3ex}
}
\caption{\textbf{(a)} Visualization of a set $V$ of sampled view axis unit vectors. \textbf{(b)} Illustration of images sampled along one view. \textbf{(c)} Illustration of multiple images sampled along multiple unique views.}
\label{fig:views}
\end{figure}

\paragraph{Isotropic Image Sampling.} 
Interpolation is needed to sample image planes not aligned with the original voxel grid. We use tri-linear and nearest-neighbour interpolation to sample the image and label map, respectively. We take advantage of the necessity for interpolation by sampling images on isotropic grids in the physical scanner space, oriented according to the patient's position in the scanner. This ensures that the model always operates on images in which the shapes of anatomical structures are maintained across scanners and acquisition protocols. Note that this approach may lead to over- or under-sampling along some axes, which may lead to loss of image information or interpolation artefacts. Empirically, however, we found that the benefit of maintaining isotropy  outweighed potential drawbacks of interpolation.

We must define a set of parameters restricting the sampling. 
Specifically, we are free to choose \textbf{1)} the pixel dimensions, $q \in \mathbb{Z^+}$ (the number of pixels to sample for each image), \textbf{2)} the real-space extent of the image (in mm), $m \in \mathbb{R^+}$, and \textbf{3)} the real-space distance between consecutive voxels, $r \in \mathbb{R^+}$. Note that two of these parameters define the third. We restrict our sampling to equal $q$, $m$ and $r$ for both image dimensions producing squared images. We sample images within a sphere of diameter $m$ centered at the origin of the scanner coordinate system. We employ a simple heuristic that attempts to pick $q$, $m$ and $r$ so that \textbf{1)} the training is computable on our GPUs with batch sizes of at least 8, \textbf{2)} $r$ approximately matches the resolution of the images along their highest resolution axis and \textbf{3)} the sampled images span the entirety of the relevant volume of all images in the dataset. When this is not possible, the requirements are prioritized in the given order, with 1 having highest priority. Note that 3 becomes less important with increasing numbers of planes as voxels missed in one plane are likely to be included in some of the others.

\paragraph{Augmentation.}
Processing the input image from different views has the  the same effect as applying affine transformations to the 3D input and presenting the transformed images to a (single-view) network. Thus, at the heart the MPUnet is a U-Net with extensive, systematic affine data augmentation. 
On top of the multi-view sampling, we also employ non-linear transformations to further augment the training data. We apply the Random Elastic Deformations algorithm \citep{RED} to each sampled image in a batch with a probability of $1/3$. The elasticity constants $\sigma$ and deformation intensity multipliers $\alpha$ are sampled uniformly from $[20, 30]$ and $[100, 500]$, respectively. This generates augmented images with high variability in terms of both deformation strength and smoothness.

The augmented images do not always display anatomically plausible structures. Yet, they often significantly improve the generalization  especially when training on small datasets or tasks involving pathologies of highly variable shape. However, we weigh the loss-contribution from augmented images by $1/3$ in order to optimize primarily over true images.

\paragraph{Pre- and post-processing.}
Our model uses a minimum of image processing outside of the network itself. We restrain from applying any post-processing of the model's output, because post-processing is typically highly task-specific. We only apply  an image- and channel-wise outlier-robust pre-possessing that scales intensity values according to the median and inter-quartile range computed over all non-background voxels. Background voxels are defined by having intensities less than or equal to the first percentile of the intensity distribution. 

\paragraph{Implementation.}
The MPUnet is available as open-source. The fully autonomous implementation makes the MPUnet applicable also for users with limited deep learning expertise and/or compute resources. A  command line interface supports fixed split or cross-validation training and evaluation on arbitrary images. Any non-constant hyperparameter can automatically be inferred from the training data. See the GitHub repository at \url{https://github.com/perslev/MultiPlanarUNet} for a user guide.

\section{Experiments and Results}
We applied the MPUNet without task-specific modifications to a total of 13 segmentation tasks. Ten of those datasets were part of the 2018 MSD challenge, described in detail and sourced on the challenge's website.
The remaining three datasets were the MICCAI 2012 Multi-Atlas Challenge (MICCAI) dataset \citep{marcus2007open}, the EADC-ADNI Harmonized Hippocampal Protocol (HarP) dataset \citep{boccardi2015training} and a knee MRI dataset from the Osteoarthritis Initiative (OAI) \cite{oai}. The evaluation covers healthy and pathological anatomical structures, mono- and multi-modal MR and CT, and various acquisition protocols.
The mean per-class F1 (dice) scores of the MPUNet are reported in Table \ref{table:results}. Note that in MSD tumour segmentation tasks 3 \& 7 both organ and tumour are segmented, and the mean F1 for those tasks is lifted by the performance on the organ and decreased by the performance on the tumour. We refer to the supplementary Table \ref{supplementary:detailed_results} for detailed per-class scores for the ten MSD tasks.

\begin{table}[t]
\centering
\caption{Performance of the MPUnet across thirteen segmentation tasks. The shown F1 (dice) scores are mean values computed across all non-background per-class F1 scores. For the 10 MSD datasets evaluation was performed by the challenge organisers on non-publicly available test-sets. For MICCAI and HarP, evaluation was performed over three  trials. Five fold cross-validation was used for OAI. The 'Classes' column include the background class, which is not included when computing the F1 scores. The 'Size' column gives the total dataset size. Note that the F1 standard deviations for tasks 8, 9 \& 10 are not yet published by the challenge organizers. We refer to \url{http://medicaldecathlon.com/results.html} for a detailed comparison of our results (team CerebriuDIKU) with those of other challenge participants.}
\renewcommand{\arraystretch}{1}
\centering
\setlength\tabcolsep{5pt}
\begin{tabular}{*{8}{l}}
&& \multicolumn{1}{l}{Dataset} & \multicolumn{1}{l}{Modality} & \multicolumn{1}{l}{Segmentation Target(s)} & \multicolumn{1}{r}{Classes} & \multicolumn{1}{r}{Size} & \multicolumn{1}{r}{F1 Score}
\\ \cmidrule[0.75pt]{3-8}
&& \multicolumn{1}{l}{MICCAI} & 
\multicolumn{1}{l}{MRI} & 
\multicolumn{1}{l}{Whole-Brain} & 
\multicolumn{1}{r}{135} & 
\multicolumn{1}{r}{35} & 
\multicolumn{1}{r}{$0.74 \pm 0.03$} \\

&& \multicolumn{1}{l}{HarP} &
\multicolumn{1}{l}{MRI} & 
\multicolumn{1}{l}{L+R Hippocampus} & 
\multicolumn{1}{r}{3} & 
\multicolumn{1}{r}{135} & 
\multicolumn{1}{r}{$0.85 \pm 0.03$} \\

&& \multicolumn{1}{l}{OAI} &
\multicolumn{1}{l}{MRI} & 
\multicolumn{1}{l}{Knee Cartilages} & 
\multicolumn{1}{r}{7} & 
\multicolumn{1}{r}{176} & 
\multicolumn{1}{r}{$0.87 \pm 0.06$} \\

\parbox[t]{0mm}{\multirow{10}{*}{\rotatebox[origin=c]{90}{2018 Medical}}} &
\parbox[t]{2mm}{\multirow{10}{*}{\rotatebox[origin=c]{90}{Segmentation Decathlon}}} &
\multicolumn{1}{|l}{Task 1} & 
\multicolumn{1}{l}{MRI} & 
\multicolumn{1}{l}{Brain Tumours} & 
\multicolumn{1}{r}{4} & 
\multicolumn{1}{r}{750} & 
\multicolumn{1}{r}{$0.60 \pm 0.24$} \\

&& \multicolumn{1}{|l}{Task 2} & 
\multicolumn{1}{l}{MRI} & 
\multicolumn{1}{l}{Cardiac, Left Atrium} & 
\multicolumn{1}{r}{2} & 
\multicolumn{1}{r}{30} & 
\multicolumn{1}{r}{$0.89 \pm 0.09$} \\

&& \multicolumn{1}{|l}{Task 3} & 
\multicolumn{1}{l}{CT} & 
\multicolumn{1}{l}{Liver \& Tumour} & 
\multicolumn{1}{r}{2} & 
\multicolumn{1}{r}{201} & 
\multicolumn{1}{r}{$0.76 \pm 0.18$} \\

&& \multicolumn{1}{|l}{Task 4} & 
\multicolumn{1}{l}{MRI} & 
\multicolumn{1}{l}{Hippocampus ROI.} & 
\multicolumn{1}{r}{2} & 
\multicolumn{1}{r}{394} & 
\multicolumn{1}{r}{$0.89 \pm 0.04$} \\

&& \multicolumn{1}{|l}{Task 5} & 
\multicolumn{1}{l}{MRI} & 
\multicolumn{1}{l}{Prostate} & 
\multicolumn{1}{r}{3} & 
\multicolumn{1}{r}{48} & 
\multicolumn{1}{r}{$0.78 \pm 0.10$} \\

&& \multicolumn{1}{|l}{Task 6} & 
\multicolumn{1}{l}{CT} & 
\multicolumn{1}{l}{Lung Tumours} & 
\multicolumn{1}{r}{2} & 
\multicolumn{1}{r}{96} & 
\multicolumn{1}{r}{$0.59 \pm 0.23$} \\

&& \multicolumn{1}{|l}{Task 7} & 
\multicolumn{1}{l}{CT} & 
\multicolumn{1}{l}{Pancreas \& Tumour} & 
\multicolumn{1}{r}{3} & 
\multicolumn{1}{r}{420} & 
\multicolumn{1}{r}{$0.48 \pm 0.21$} \\

&& \multicolumn{1}{|l}{Task 8} & 
\multicolumn{1}{l}{CT} & 
\multicolumn{1}{l}{Hepatic Ves. \& Tumour} & 
\multicolumn{1}{r}{3} & 
\multicolumn{1}{r}{443} & 
\multicolumn{1}{l}{$0.49$} \\

&& \multicolumn{1}{|l}{Task 9} & 
\multicolumn{1}{l}{CT} & 
\multicolumn{1}{l}{Spleen} & 
\multicolumn{1}{r}{2} & 
\multicolumn{1}{r}{61} & 
\multicolumn{1}{l}{$0.95$} \\

&& \multicolumn{1}{|l}{Task 10} & 
\multicolumn{1}{l}{CT} & 
\multicolumn{1}{l}{Colon Cancer} & 
\multicolumn{1}{r}{2} & 
\multicolumn{1}{r}{190} & 
\multicolumn{1}{l}{$0.28$}

\end{tabular}
\label{table:results}
\end{table}

The MPUnet reached state-of-the-art performance for DL methods on the three non-challenge datasets (MICCAI, HaRP and OAI) despite comparable methods being developed and tuned specifically to the cohorts and tasks. On MICCAI, with a mean F1 of 0.74 the MPUnet compares similar to the 0.74 obtained in \cite{miccai_1} using a 2D multi-scale CNN on brain-extracted images and 0.75 obtained in \cite{miccai_2} using a combination of a multi-scale 2D CNN, 3D patch-based CNN, a spatial information encoder network and a probabilistic atlas also on brain-extracted images. With a mean F1 of 0.85 on HarP, the MPUnet compares favorable to 0.78-0.83 (depending on subject disease state) reported in \cite{quickNAT}. On OAI, with a mean F1 of 0.87, the MPUnet gets near the 0.88/0.89 (baseline/follow-up) obtained in \cite{ambellan} using a task-specific pipeline including 2D- and 3D U-nets along with multiple statistical shape model refinement steps. However,  the comparison cannot be directly made as \cite{ambellan} worked on a smaller subset of the OAI data and predicted only 4 classes while we distinguished 7.

The MPUnet ranked 5th and 6th place in the first and second phases of the Medical Segmentation Decathlon respectively, in most cases comparing unfavorable only to significantly more compute intensive systems (see  below).\footnote{For comparison, the median F1 scores over all 10 tasks of the best five phase 1 submissions were
0.74, 0.67, 0.69, 0.66, and (our method) 0.69. Note that the official ranking was based on a more rigorous statistical analysis.}

The question arises how the performance of a 2D U-net with multi-planar augmentation compares to a U-net with 3D convolutions.
Such 3D models are computationally demanding and typically need -- in our experience -- large training datasets to achieve proper generalization. While we are not making the claim that the MPUnet is universally superior to 3D models, we did find the MPUnet to outperform a 3D U-net of comparable topology, learning and augmentation procedure across multiple tasks including one for which the 3D model had sufficient spatial extent to operate on the entire input volume at once. We refer to the supplementary Table~\ref{supplementary:3d_comp} for details. We also found the MPUnet superior to both single 2D U-Nets trained on individual planes as well as ensembles of separate 2D U-Nets trained on different planes, see Table~\ref{supplementary:number_of_planes} \& \ref{supplementary:mpu_ensemble} and  Fig.~\ref{supplementary:smmv_vs_mmmv} in the supplementary material.

\section{Discussion and Conclusions}
The  empirical evaluation over 13 segmentation tasks showed that multi-planar augmentation provides a simple mechanism for obtaining accurate segmentation models without hyperparameter tuning.
With no task-specific modifications the MPUnet performs well across many non-pathological tissues imaged with various MR and CT protocols, in spite of the target compartments varying drastically in  number, physical size, shape- and spatial distributions, as well as contrast to the surrounding tissues. Also the accuracies on the more difficult pathological targets are favorable compared to most other MSD contesters.

The MSD winning algorithm \citep{insee} relied on selecting a suitable model topology and/or cascade from an ensemble of candidates through cross-validation. In contrast to this and other top-ranking participants, we were interested to develop a task-agnostic segmentation system based on a single architecture and learning procedure that makes the system lightweight and easily transferable to clinical settings with limited compute resources.

That the MPUnet can be applied 'as is' across many tasks with high performance 
and its robustness against overfitting can be attributed to both the fully convolutional network approach, which is already known to generalize well, and  our multi-planar augmentation framework. The latter allows us to apply a single 2D model with fixed hyperparameters, resulting in a fully autonomous segmentation system of low computational complexity. Multi-planar training improves the generalization performance in several ways: \textbf{1)} Sampling from multiple planes allows for a huge number of anatomically relevant images augmenting the training data; \textbf{2)} Exposing a 2D model to multiple planes takes the 3D nature of the input into account while maintaining the statistical and computational efficiency of 2D kernels; \textbf{3)} The systematic augmentation scheme allows test time augmentation to be performed, which increases the performance through variance reduction if errors across views are uncorrelated for a given subject (visualized in supplementary Fig.~\ref{supplementary:prob_vis}).
This makes the MPUnet an open source alternative to 3D fully convolutional neural networks.

\section*{Acknowledgements}
We would like to thank both Microsoft and NVIDIA for providing computational resources on the Azure platform for this project.

\bibliographystyle{class_files/splncs}
\bibliography{bibliography/bibliography}

\newpage
\appendix
\renewcommand\thefigure{S.\arabic{figure}}
\renewcommand\thetable{S.\arabic{table}}
\section*{Supplementary Material}
\setcounter{figure}{0}
\setcounter{table}{0}

\begin{figure}[h]
	\centering
    \makebox[\textwidth][c]{
    	\includegraphics[width=1.0\linewidth]{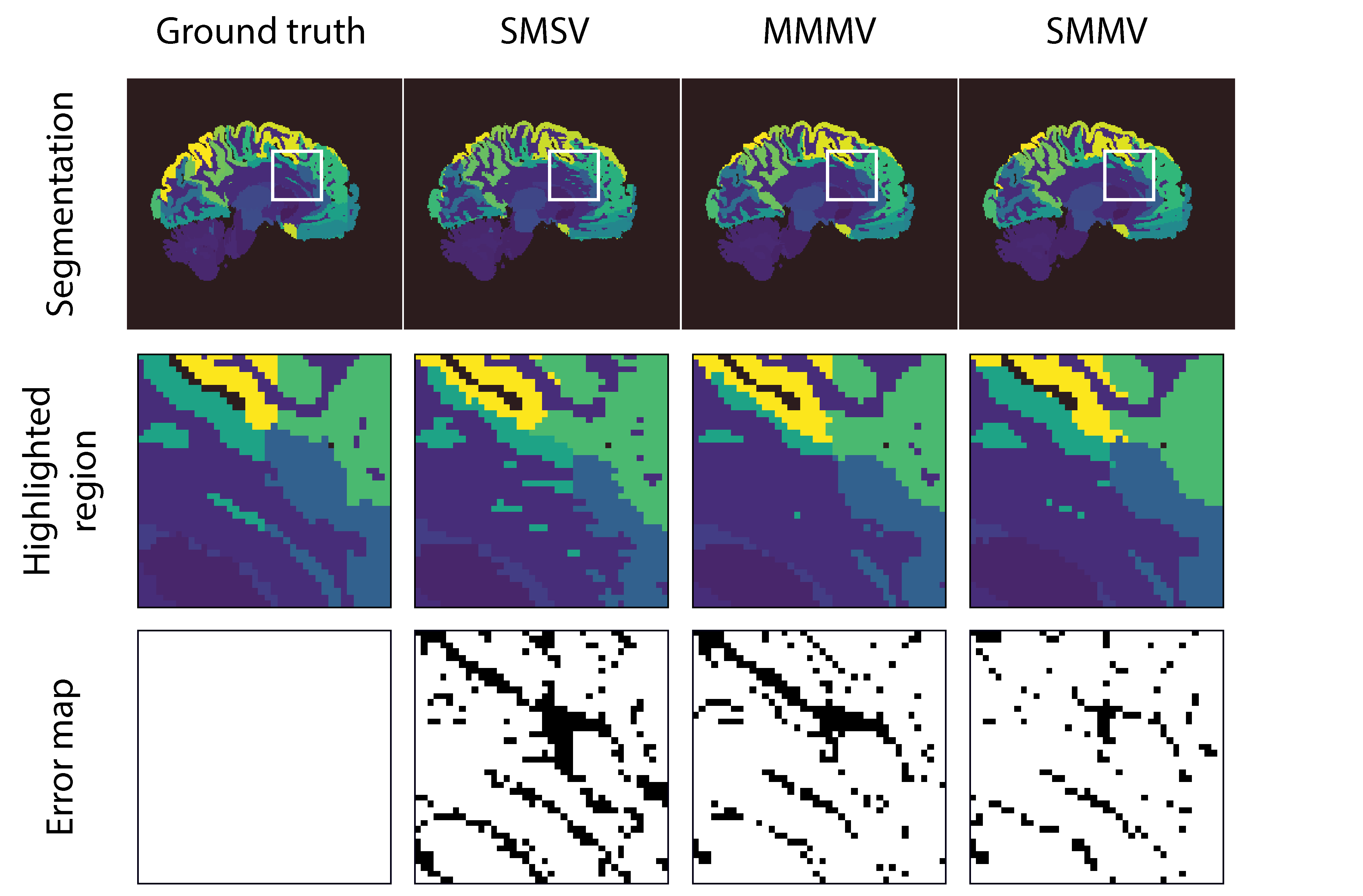}
    }
    \caption{Visual comparison of the typical performance improvements obtained on a random subject of the MICCAI dataset when going from a single U-Net model fit to a single plane (single-model-single-view, SMSV, second column) to an ensemble of such models (multi-model-multi-view, MMMV, third column) to the MPUnet (single-model-multi-view, SMMV, fourth column). The first row shows the full segmentation on a single 2D slice. The second row presents a zoom of the highlighted region shown in each image of row 1. The third row shows a binary error-map for the highlighted region with black pixels representing errors compared to the ground truth and white pixels representing correctly classified pixels.}
    \label{supplementary:smmv_vs_mmmv}
\end{figure}

\begin{figure}[h]
	\centering
    \makebox[\textwidth][c]{
    	\includegraphics[width=1.0\linewidth]{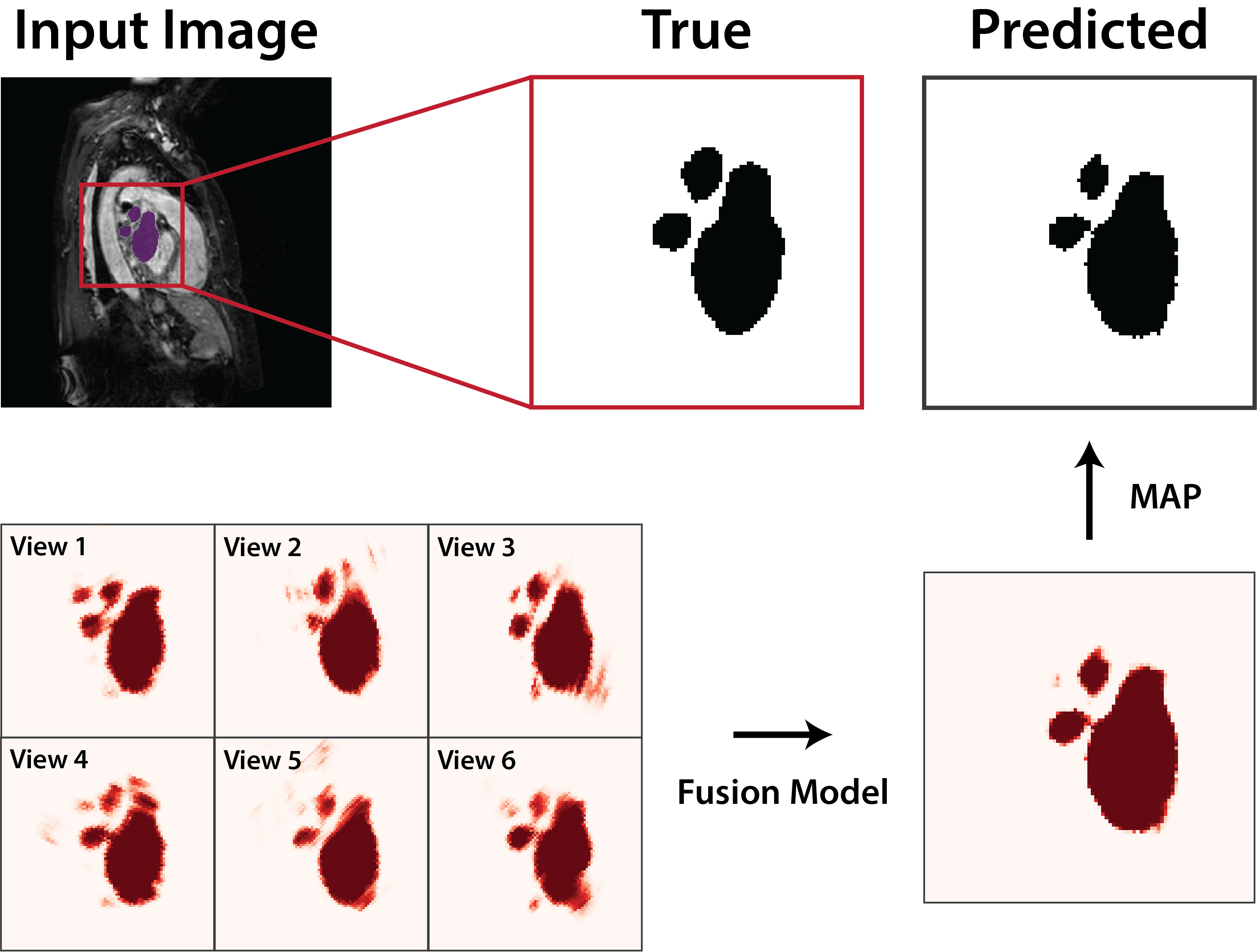}
    }
    \caption{Visualization of the benefit of the MPUNet test-time augmentation approach. A 2D slice from an input image is shown in the upper left panel with a highlighted region of interest to the right giving the ground truth (binary) label map for the left atrium of an image in the Medical Segmentation Decathlon Task 4 dataset. A single MPUnet predicts on the entire image volume along 6 planes and maps the predictions to the input image space, producing a set of 6 segmentation volumes. For each of those, the corresponding slice to the input image is shown in the lower left panel. Darker red colors indicate higher confidence of the model in the foreground class at the given pixel as seen in a given view. Note that while each confidence map matches the ground truth to a large extend, the model has both false positive and false negative confidence in certain areas of individual views. After passing the 6 segmentation maps through the fusion model (lower right), a much cleaner output is produced, which after thresholding (upper right) coresponds well to the ground truth.}
    \label{supplementary:prob_vis}
\end{figure}

\begin{table}
\centering
\caption{Fixed hyperparameter set for the optimization of the MPUnet core model on any segmentation task.}
\renewcommand{\arraystretch}{1}
\centering
\setlength\tabcolsep{7pt}
\begin{tabular}{*{3}{l}}
\toprule
\multicolumn{1}{r}{Parameter} & \multicolumn{1}{c}{Value} & \multicolumn{1}{c}{Notes} \\ 
\midrule
\multicolumn{1}{r}{Optimizer} & \multicolumn{1}{l}{Adam} &
\multirow[t]{5}{155pt}{The global learning rate is reduced by 10 \% for every 2 consecutive epochs without validation performance improvements.} \\
\multicolumn{1}{r}{\textit{Learning rate} -} & \multicolumn{1}{l}{$5\cdot 10^{-5}$} & \\
\multicolumn{1}{r}{$\beta_1$ -} & \multicolumn{1}{l}{$0.9$} & \\
\multicolumn{1}{r}{$\beta_2$ -} & \multicolumn{1}{l}{$0.999$} & \\
\multicolumn{1}{r}{$\epsilon$ -} & \multicolumn{1}{l}{$1\cdot 10^{-8}$} & \\
\midrule
\multicolumn{1}{r}{Loss function} & \multicolumn{1}{l}{Cross entropy} &
\multirow[t]{1}{155pt}{} \\
\multicolumn{1}{r}{\textit{Regularization} -} & \multicolumn{1}{l}{None} & \\
\multicolumn{1}{r}{\textit{Class balancing} -} & \multicolumn{1}{l}{None} & \\
\midrule
\multicolumn{1}{r}{Model Topology} & \multicolumn{1}{l}{2D U-Net} &
\multirow[t]{4}{155pt}{The input dimensions are inferred based on the sizes of the images of the training data cohort. The range of 128-512 is appropriate for typical compute systems, but may be expanded to work on larger images. Generalization properties outside of this suggested range have not been tested. Note that small images volumes may be oversampled.} \\
\multicolumn{1}{r}{\textit{Input dim} -} & \multicolumn{1}{l}{128-512} & \\
\multicolumn{1}{r}{\textit{Depth} -} & \multicolumn{1}{l}{4} & \\
\multicolumn{1}{r}{\textit{Up-sampling} -} & \multicolumn{1}{l}{Nearest neighbour} & \\
\multicolumn{1}{r}{\textit{Activations} -} & \multicolumn{1}{l}{ReLU} & \\
\multicolumn{1}{r}{\textit{Conv. kernel size} -} & \multicolumn{1}{l}{$3 \times 3$} & \\
\multicolumn{1}{r}{\textit{Max-pool kernel size} -} & \multicolumn{1}{l}{$2 \times 2$} & \\
\multicolumn{1}{r}{\textit{Padding} -} & \multicolumn{1}{l}{True ('same')} & \\
\multicolumn{1}{r}{\textit{Batch normalization} -} & \multicolumn{1}{l}{True} & \\
\multicolumn{1}{r}{\textit{Parameters} -} & \multicolumn{1}{l}{$6.2\cdot10^{7}$} & \\
\midrule
\multicolumn{1}{r}{Image sampling} & \multicolumn{1}{l}{Multi-Planar} &
\multirow[t]{3}{155pt}{Plane unit vectors are sampled uniformly from the 3-sphere with at least $60\deg$ angle between them.} \\
\multicolumn{1}{r}{\textit{Image interp} -} & \multicolumn{1}{l}{Tri-linear} & \\
\multicolumn{1}{r}{\textit{Label interp} -} & \multicolumn{1}{l}{Nearest-neighbour} & \\
\multicolumn{1}{r}{\textit{Num. planes} -} & \multicolumn{1}{l}{6} & \\
\midrule
\multicolumn{1}{r}{Non-linear aug.} & \multicolumn{1}{l}{RED*} &
\multirow[t]{3}{155pt}{Strength and smoothness sampled on-the-fly to produce variable deformations. *Random Elastic Deformations.} \\
\multicolumn{1}{r}{\textit{Strength, $\alpha$} -} & \multicolumn{1}{l}{uniform(100, 500)} & \\
\multicolumn{1}{r}{\textit{Elasticity, $\sigma$} -} & \multicolumn{1}{l}{uniform(20, 30)} & \\
\multicolumn{1}{r}{\textit{Apply prob.} -} & \multicolumn{1}{l}{$1/3$} & \\
\multicolumn{1}{r}{\textit{Loss weight} -} & \multicolumn{1}{l}{$1/3$} & \\
\midrule
\multicolumn{1}{r}{Pre-processing} & \multicolumn{1}{l}{Robust scaling} &
\multirow[t]{3}{155pt}{Image- and channel-wise scaling to (non-background) intensity distribution of median 0 and IQR 1.} \\
\multicolumn{1}{r}{Post-processing} & \multicolumn{1}{l}{None} & \\
&& \\
\midrule
\multicolumn{1}{r}{Batch size} & \multicolumn{1}{l}{8-16} &
\multirow[t]{5}{155pt}{16 by default, reduced by 2 until batches fit in GPU memory. A fraction of 1 minus the mean validation recall of a batch must contain non-background images ($\geq 1$ pixel of class $\neq 0$).} \\
\multicolumn{1}{r}{\textit{Foreground fraction} -} & \multicolumn{1}{l}{1 - recall} & \\
&& \\ && \\ && \\ && \\
\midrule
\multicolumn{1}{r}{Training epochs} & \multicolumn{1}{l}{$\infty$} &
\multirow[t]{3}{155pt}{Training continues until 15 consecutive epochs of without validation performance improvements.} \\
\multicolumn{1}{r}{\textit{Train images/epoch -}} & \multicolumn{1}{l}{2500} & \\
\multicolumn{1}{r}{\textit{Val. images/epoch -}} & \multicolumn{1}{l}{3500} & \\
\midrule
\multicolumn{1}{r}{Early stopping criteria} & \multicolumn{1}{l}{Validation F1} &
\multirow[t]{3}{155pt}{Mean per-class F1 scores (excluding background) computed over all images of a validation epoch.} \\
\multicolumn{1}{r}{Model selection criteria} & \multicolumn{1}{l}{Validation F1} & \\
&& \\
\bottomrule
\label{supplementary:hyperparams}
\end{tabular}
\end{table}

\begin{table}
\centering
\caption{F1 improvement on the MICCAI and MSD Task 4 datasets for a MPUnet of 2-9 planes relative to the mean performance of 9 single-plane models each fit to 1 of the 9 planes of the 9-plane MPUnet model. While the absolute performance benefit of using higher numbers of planes vary between the two tasks, the gains are monotonically increasing with views across both. Note that these results are only guiding as the experiments were conducted just once for each MPUnet.}
\renewcommand{\arraystretch}{1}
\centering
\setlength\tabcolsep{6pt}
\begin{tabular}{*{10}{l}}
\multicolumn{1}{r}{Num. planes, $i=$} & \multicolumn{1}{c}{9} & \multicolumn{1}{c}{8} & \multicolumn{1}{c}{7} & \multicolumn{1}{c}{6} & \multicolumn{1}{c}{5} & \multicolumn{1}{c}{4} & \multicolumn{1}{c}{3} & \multicolumn{1}{c}{2}
\\ \midrule
\multicolumn{1}{r}{MICCAI} & \multicolumn{1}{c}{$0.041$} & \multicolumn{1}{c}{$0.037$} & \multicolumn{1}{c}{$0.037$} & \multicolumn{1}{c}{$0.035$} & \multicolumn{1}{c}{$0.029$} & \multicolumn{1}{c}{$0.024$} & \multicolumn{1}{c}{$0.015$} & \multicolumn{1}{c}{$0.012$} \\
\multicolumn{1}{r}{MSD T4} & \multicolumn{1}{c}{$0.017$} & \multicolumn{1}{c}{$0.017$} & \multicolumn{1}{c}{$0.016$} & \multicolumn{1}{c}{$0.015$} & \multicolumn{1}{c}{$0.015$} & \multicolumn{1}{c}{$0.014$} & \multicolumn{1}{c}{$0.013$} & \multicolumn{1}{c}{$0.012$} \\
\midrule
\label{supplementary:number_of_planes}
\end{tabular}
\end{table}

\begin{table}
\centering
\caption{Mean F1 performance on the MICCAI dataset for MPUnets of $i \in \{3, 6, 9 \}$ planes compared to ensembles of individual single-plane model each trained on a unique plane. Each single-plane model is optimized under the same set of hyperparameter as the MPUnet. Note that the single-planar ensembles have $i$ times the parameters of their MPUnet counterparts divided evenly across its $i$ sub-models.}
\renewcommand{\arraystretch}{1}
\centering
\setlength\tabcolsep{8pt}
\begin{tabular}{*{10}{l}}
\multicolumn{1}{r}{Num. planes, $i=$} & \multicolumn{1}{c}{9} & \multicolumn{1}{c}{6} & \multicolumn{1}{c}{3}
\\ \midrule
\multicolumn{1}{r}{Single-Planar Ensemble} & \multicolumn{1}{c}{$0.717 \pm 0.019$} & \multicolumn{1}{c}{$0.714 \pm 0.021$} & \multicolumn{1}{c}{$0.710 \pm 0.024$} \\
\multicolumn{1}{r}{Multi-Planar U-Net} & \multicolumn{1}{c}{$0.743 \pm 0.028$} & \multicolumn{1}{c}{$0.737 \pm 0.027$} & \multicolumn{1}{c}{$0.717 \pm 0.030$} \\
\midrule
\label{supplementary:mpu_ensemble}
\end{tabular}
\end{table}

\begin{table}[h]
\centering
\caption{Detailed report of the MPUnet mean and standard deviation F1 (dice) performance on individual target classes across the 10 tasks of the Medical Segmentation Decathlon.}
\renewcommand{\arraystretch}{1}
\centering
\setlength\tabcolsep{5pt}
\begin{tabular}{*{4}{l}}
\multicolumn{1}{l}{Dataset} & \multicolumn{1}{l}{Description} & \multicolumn{1}{l}{Class} & \multicolumn{1}{r}{F1 Score} \\
\midrule
\multicolumn{1}{l}{Task 1} & 
\multicolumn{1}{l}{Brain Tumours} & 
\multicolumn{1}{l}{Edema} & \multicolumn{1}{r}{$0.70 \pm 0.20$} \\ && 
\multicolumn{1}{l}{Non-enhancing tumor} & \multicolumn{1}{r}{$0.43 \pm 0.31$} \\ &&
\multicolumn{1}{l}{Enhancing tumour} & \multicolumn{1}{r}{$0.67 \pm 0.22$} \\

\multicolumn{1}{l}{Task 2} & 
\multicolumn{1}{l}{Cardiac} & 
\multicolumn{1}{l}{Left atrium} & \multicolumn{1}{r}{$0.89 \pm 0.09$} \\

\multicolumn{1}{l}{Task 3} & 
\multicolumn{1}{l}{Liver \& Tumour} & 
\multicolumn{1}{l}{Liver} & \multicolumn{1}{r}{$0.94 \pm 0.03$} \\ &&
\multicolumn{1}{l}{Cancer} & \multicolumn{1}{r}{$0.57 \pm 0.32$} \\

\multicolumn{1}{l}{Task 4} & 
\multicolumn{1}{l}{Hippocampus ROI.} & 
\multicolumn{1}{l}{Anterior} & \multicolumn{1}{r}{$0.90 \pm 0.03$} \\ &&
\multicolumn{1}{l}{Posterior} & \multicolumn{1}{r}{$0.88 \pm 0.04$} \\

\multicolumn{1}{l}{Task 5} & 
\multicolumn{1}{l}{Prostate} & 
\multicolumn{1}{l}{Peripheral zone} & \multicolumn{1}{r}{$0.69 \pm 0.13$} \\ &&
\multicolumn{1}{l}{Transition zone} & \multicolumn{1}{r}{$0.86 \pm 0.07$} \\

\multicolumn{1}{l}{Task 6} & 
\multicolumn{1}{l}{Lung Tumours} & 
\multicolumn{1}{l}{Cancer} & \multicolumn{1}{r}{$0.59 \pm 0.23$} \\

\multicolumn{1}{l}{Task 7} & 
\multicolumn{1}{l}{Pancreas \& Tumour} & 
\multicolumn{1}{l}{Pancreas} & \multicolumn{1}{r}{$0.71 \pm 0.14$} \\ &&
\multicolumn{1}{l}{Cancer} & \multicolumn{1}{r}{$0.25 \pm 0.27$} \\

\multicolumn{1}{l}{Task 8} & 
\multicolumn{1}{l}{Hepatic Ves. \& Tumour} & 
\multicolumn{1}{l}{Vessel} & \multicolumn{1}{l}{$0.59$} \\ &&
\multicolumn{1}{l}{Tumour} & \multicolumn{1}{l}{$0.38$} \\

\multicolumn{1}{l}{Task 9} & 
\multicolumn{1}{l}{Spleen} & 
\multicolumn{1}{l}{Spleen} & \multicolumn{1}{l}{$0.95$} \\

\multicolumn{1}{l}{Task 10} & 
\multicolumn{1}{l}{Colon Cancer} & 
\multicolumn{1}{l}{Cancer primaries} & \multicolumn{1}{l}{$0.28$} \\

\bottomrule
\end{tabular}
\label{supplementary:detailed_results}
\end{table}

\begin{table}
\centering
\caption{Comparison of the Multi-Planar UNet and a 3D UNet of identical topology (all 2D operations replaced by 3D operations) on the three non-challenge benchmark datasets MICCAI, HaRP and OAI as well as the Medical Segmentation Decathlon (MSD) Task 4 dataset (hippocampus in region-of-interest). The two models were trained under identical optimization parameters. The shown scores are mean per-class F1 scores pooled across three separate training and evaluation sessions. The MSD Task 4 dataset experiments were conducted on random splits of the challenge training data, as we do not have access to the test set. The 3D UNet was trained on isotropic ROIs of 64-cube voxels with random rotations and 3D random elastic deformations applied at batch-sampling time. This was done to emulate the benefit of the MPUNet's significant data augmentation. The sampled voxel-resolution was identical to that chosen for the MPUNet. The 3D model has a total of $90$ million parameters against the $62$ of the MPUnet. The MSD Task 4 dataset consists of small cut-out regions of interest spanning narrowly around the hippocampus to segment, and was include here to study the performance of the 3D model when the entire input image fits within the 64-cube input patch.
\textbf{Note:} The OAI dataset used for those experiments was a smaller subset of the full dataset for which results are displayed in Table \ref{table:results} (no follow-up scans included, specifically).}
\renewcommand{\arraystretch}{1}
\centering
\setlength\tabcolsep{7pt}
\begin{tabular}{*{5}{l}}
\multicolumn{1}{r}{} & \multicolumn{1}{c}{MICCAI} & \multicolumn{1}{c}{HaRP} & \multicolumn{1}{c}{OAI} & \multicolumn{1}{c}{MSD T4}
\\ \midrule
\multicolumn{1}{r}{3D U-Net w. rotations} & \multicolumn{1}{c}{$0.74 \pm 0.04$} & \multicolumn{1}{c}{$0.84 \pm 0.05$} & \multicolumn{1}{c}{$0.81 \pm 0.07$} & \multicolumn{1}{c}{$0.87 \pm 0.04$} \\
\multicolumn{1}{r}{Multi-Planar U-Net} & \multicolumn{1}{c}{$0.74 \pm 0.03$} & \multicolumn{1}{c}{$0.85 \pm 0.03$} & \multicolumn{1}{c}{$0.84 \pm 0.07$} & \multicolumn{1}{c}{$0.88 \pm 0.04$} \\
\midrule
\label{supplementary:3d_comp}
\end{tabular}
\end{table}

\begin{table}[]
\begin{center}
\caption{MPUnet base model topology (U-Net type) for images sampled with pixel dim $q=256$. Note: Convolution strides of $1\times1$ where used in all layers.}
\setlength\tabcolsep{5pt}
\begin{tabular}{r|ccccc}
\textbf{Layer name} & \textbf{Output dim} & \textbf{Kernel dim} & \textbf{Filters} & \textbf{Activation} & \textbf{Pad} \\
\toprule
Input               & $256\times256\times C$          & -                   & -                & -                   & -            \\
conv\_1\_1          & $256\times256\times90$          & $3\times3$                 & 90               & ReLU                & same         \\
conv\_1\_2          & $256\times256\times90$          & $3\times3$                 & 90               & ReLU                & same         \\
bn\_1               & $256\times256\times90$          & -                   & -                & -                   & -            \\
pool\_1             & $128\times128\times90$            & $2\times2$                 & -                & -                   & valid        \\
conv\_2\_1          & $128\times128\times181$           & $3\times3$                 & 181              & ReLU                & same         \\
conv\_2\_2          & $128\times128\times181$           & $3\times3$                 & 181              & ReLU                & same         \\
bn\_2               & $128\times128\times181$           & -                   & -                & -                   & -            \\
pool\_2             & $64\times64\times181$          & $2\times2$                 & -                & -                   & valid        \\
conv\_3\_1          & $64\times64\times362$           & $3\times3$                 & 362              & ReLU                & same         \\
conv\_3\_2          & $64\times64\times362$           & $3\times3$                 & 362              & ReLU                & same         \\
bn\_3               & $64\times64\times362$           & -                   & -                & -                   & -            \\
pool\_3             & $32\times32\times362$           & $2\times2$                 & -                & -                   & valid        \\
conv\_4\_1          & $32\times32\times724$           & $3\times3$                 & 724              & ReLU                & same         \\
conv\_4\_2          & $32\times32\times724$           & $3\times3$                 & 724              & ReLU                & same         \\
bn\_4               & $32\times32\times724$           & -                   & -                & -                   & -            \\
pool\_4             & $16\times16\times724$           & $2\times2$                 & -                & -                   & valid        \\
conv\_5\_1          & $16\times16\times1448$          & $3\times3$                 & 1448             & ReLU                & same         \\
conv\_5\_2          & $16\times16\times1448$          & $3\times3$                 & 1448             & ReLU                & same         \\
up\_1               & $32\times32\times1448$          & $2\times2$                 & -                & -                   & -            \\
conv\_6\_0             & $32\times32\times724$           & $2\times2$                 & 724              & ReLU                & same         \\
bn\_6            & $32\times32\times724$           & -                   & -                & -                   & -            \\
merge(bn4, bn6)     & $32\times32\times1448$          & -                   & -                & -                   & -            \\
conv\_6\_1          & $32\times32\times724$           & $3\times3$                 & 724              & ReLU                & same         \\
conv\_6\_2          & $32\times32\times724$           & $3\times3$                 & 724              & ReLU                & same         \\
bn\_7               & $32\times32\times724$           & -                   & -                & -                   & -            \\
up\_2               & $64\times64\times724$           & $2\times2$                 & -                & -                   & -            \\
conv\_7\_0          & $64\times64\times362$           & $2\times2$                 & 362              & ReLU                & same         \\
bn\_8               & $64\times64\times362$           & -                   & -                & -                   & -            \\
merge(bn3, bn8)     & $64\times64\times724$           & -                   & -                & -                   & -            \\
conv\_7\_1          & $64\times64\times362$           & $3\times3$                 & 362              & ReLU                & same         \\
conv\_7\_2          & $64\times64\times362$           & $3\times3$                 & 362              & ReLU                & same         \\
bn\_9               & $64\times64\times362$                   & -                   & -                & -                   & -            \\
up\_3               & $128\times128\times362$           & $2\times2$                 & -                & -                   & -            \\
conv\_8\_0          & $128\times128\times181$           & $2\times2$                 & 181              & ReLU                & same         \\
bn\_10              & $128\times128\times181$           & -                   & -                & -                   & -            \\
merge(bn2, bn10)    & $128\times128\times362$           & -                   & -                & -                   & -            \\
conv\_8\_1          & $128\times128\times181$           & $3\times3$                 & 181              & ReLU                & same         \\
conv\_8\_2          & $128\times128\times181$           & $3\times3$                 & 181              & ReLU                & same         \\
bn\_11              & $128\times128\times181$           & -                   & -                & -                   & -            \\
up\_4               & $256\times256\times181$         & $2\times2$                 & -                & -                   & -            \\
conv\_9\_0          & $256\times256\times90$          & $2\times2$                 & 90               & ReLU                & same         \\
bn\_12              & $256\times256\times90$          & -                   & -                & -                   & -            \\
merge(bn1, bn12)    & $256\times256\times180$         & -                   & -                & -                   & -            \\
conv\_9\_1          & $256\times256\times90$          & $3\times3$                & 90               & ReLU                & same         \\
conv\_9\_2          & $256\times256\times90$          & $3\times3$                 & 90               & ReLU                & same         \\
bn\_13              & $256\times256\times90$          & -                   & -                & -                   & -            \\
output              & $256\times256\times K$         & $1\times1$                 & K              & softmax             & -            \\
\bottomrule
\end{tabular}
\text{}\\
\textbf{Trainable parameters: $62,062,342$ (for $K=135$, $C=1$)}
\label{supplementary:topology}
\end{center}
\end{table}

\end{document}